\documentclass[12pt]{article}

\usepackage{palatino}
\usepackage{wrapfig,graphicx,psfrag,xspace,color,latexsym,amsmath,amssymb,booktabs}
\usepackage{comment}
\usepackage{setspace}
\usepackage[dvips]{epsfig}      % PostScript figures
\usepackage{longtable}          % multipage tables
\usepackage{url,hyperref}       % url?
\usepackage{multirow}           % Multirow table entries
\usepackage{color}
\usepackage{algorithm}
\usepackage{algorithmic}
\usepackage{graphicx}
\usepackage{tikz}
\usepackage{xcolor}
\usepackage{euscript,enumerate,float,afterpage}
\usepackage{rotate}
\usepackage{verbatim}
\usepackage{epstopdf}
\usepackage{hhline}
\usepackage{subcaption}
\usepackage[font=footnotesize,labelfont=bf]{caption}
\usepackage[utf8]{inputenc}
\usepackage{listings}
\usepackage{xcolor}
\usepackage{makecell}

\definecolor{codegreen}{rgb}{0,0.5,0}
\definecolor{codegray}{rgb}{0.5,0.5,0.5}
\definecolor{codepurple}{rgb}{0.58,0,0.82}
\definecolor{backcolour}{rgb}{0.95,0.95,0.92}
\lstdefinestyle{mystyle}{
  backgroundcolor=\color{backcolour},   commentstyle=\color{codegreen},
  keywordstyle=\color{magenta},
  numberstyle=\tiny\color{codegray},
  stringstyle=\color{codepurple},
  basicstyle=\ttfamily\large,
  breakatwhitespace=false,         
  breaklines=true,                 
  captionpos=b,                    
  keepspaces=true,                 
  numbers=left,                    
  numbersep=5pt,                  
  showspaces=false,                
  showstringspaces=false,
  showtabs=false,                  
  tabsize=2,morekeywords={predict,detect},
    keywordstyle=\bfseries,
    showstringspaces=false
}

\def\cf{{\it cf.}~}
\def\ie{{\it i.e.},~}
\def\eg{{\it e.g.},~}

\newcommand{\squishlist}{
   \begin{list}{$\bullet$}
    { \setlength{\itemsep}{0pt}      \setlength{\parsep}{0pt}
      \setlength{\topsep}{3pt}       \setlength{\partopsep}{0pt}
      \setlength{\listparindent}{-2pt}
      \setlength{\itemindent}{-5pt}
      \setlength{\leftmargin}{1em} \setlength{\labelwidth}{0em}
      \setlength{\labelsep}{0.5em} } }

\newcommand{\squishend}{
    \end{list}  }

\begin{document}

\title{Backdoor Attack and Defense for Deep Regression}

\author{
Xi L$\mbox{i}^*$, George Kesidi$\mbox{s}^{*\dagger}$, 
David J. Mille$\mbox{r}^{*\dagger}$,  Vladimir Luci$\mbox{c}^{\ddagger\dagger}$\\
\{xzl45,gik2,djm25\}@psu.edu,  v.lucic@imperial.ac.uk \\
\begin{tabular}{ccc}
$\mbox{ }^*$Pennsylvania State University & $\mbox{ }^\dagger$Anomalee Inc. & 
$\mbox{ }^\ddagger$Imperial College
\end{tabular}
}

\date{}

\maketitle

\begin{abstract}
We demonstrate a backdoor attack on deep regression, with an example from financial derivatives pricing.
The backdoor attack is localized based on training-set data poisoning 
wherein the mislabeled samples are surrounded by 
correctly supervised ones. We demonstrate how such localization is 
necessary for attack success. 
We also study the performance of a backdoor defense
using gradient-based discovery of local error maximizers.
Local error maximizers which are associated with significant
(interpolation) error, and are
proximal to many training samples, 
are suspicious.
This method is also used to accurately train for deep regression in the first
place by active (deep) learning leveraging an ``oracle"
capable of providing 
real-valued supervision (a regression target) for samples.
Such oracles, including traditional numerical solvers of the pricing PDEs are far more computationally
costly at (test-time) inference, compared to deep regression.
\end{abstract}

\section{Introduction}

Deep regression \cite{LMPH19},
active learning for regression \cite{TPS16,Kading18,WLH18}, 
and deep active learning \cite{Ren20}
 have received significant interest over the past
few years.
We assume an automated ``oracle" $Z$ 
capable of providing 
real-valued supervision (a regression target) for samples
is available for purposes of training. 
A parameterized model $Y$ for regression, which could be a deep neural network (DNN),
is typically trained to approximate the
oracle based on the mean squared error (MSE) training objective, \ie
% \eg the mean normalized square error,
% \begin{align} \label{MSE0}
%   \frac{1}{|S|} \sum_{x\in S} \left|\frac{Y(x)-Z(x)}{Z(x)}\right|^2
% \end{align}
\begin{align} \label{MSE0}
  \frac{1}{|S|} \sum_{x\in S} \left|Y(x)-Z(x)\right|^2
\end{align}
where $S$ is the training set of supervised input samples $x$, with $Z(x)$ the supervising target for $x$.
The benefit of a
DNN model is that it can perform inference at much higher speed 
than the oracle \cite{CC21}.
This benefit at test/production time
needs to greatly outweigh the cost of invoking
the oracle during training.
Moreover, a DNN is preferable to simpler models
when the oracle is a complicated (highly nonconvex) function of 
the input and the inputs belong to a 
high-dimensional sample space.
In \cite{XKM21}, we described a deep active learning approach
for regression in this context.
The active learning employs a gradient based method to 
find local maximizers of the error between $Y$ and $Z$
which are then used for purposes of retraining,
\cf Section \ref{sec:method}.

In Section \ref{sec:attack}, we demonstrate how a backdoor can
be planted in a DNN used for regression via training-set
data poisoning \cite{backdoor-imperceptible,backdoor-perceptible,RE20}.
As both an illustrative
example and a real application, we consider an oracle which evaluates
a single-barrier (simple) financial option.
In Section \ref{sec:defense}, we show how the 
method of Section \ref{sec:method} can be leveraged to detect
and mitigate the backdoor.
%The paper concludes with a discussion including future work in
%Section \ref{sec:fw}.

\section{Gradient-Based Method for 
Local Error Maximizers \cite{XKM21}}\label{sec:method}

Let $Y$ be the trained DNN regression model and $Z$ be the
oracle.
A gradient-ascent sequence with index $n=0,1,2,...$, seeking a local maximizer of the square error ${\mathcal E}=|Y-Z|^2$, is: 
\begin{equation}\label{GA}
    x_{n+1} = x_n + s_n\nabla{\mathcal E}(x_n),
\end{equation}
where $x_0$ is the initialization, the step size $s_n$ is non-increasing with $n$, a
finite-difference approximation is used for the gradient of the oracle $Z$, and the gradient of the neural network model $Y$ with respect to the
input is computed by back-propagation \cite{Hwang97}.

This process is iteratively repeated
in the manner of classical iterated gradient based
hill-climbing  exploration/exploitation optimization 
methods, more recently called
Cuckoo search \cite{YD09}:
At step $k+1$,  gradient ascent search  
of the error $|Y-Z|^2$ is seeded with the 
current local maximizers with largest absolute errors 
%(denoted $S_k$, 
(obtained from the $k$-th step),
and uses smaller initial step size and
tighter stopping condition than for step $k$.

Many obvious variations of the foregoing approach are possible.
Though we assume the frequent use of the
oracle is justified for training,
computational costs of training could be much more significant for
more complex oracles, for instance those that numerically price exotic, path-dependent options (see, e.g. \cite{Bouzoubaa10} for examples of
such contracts; a particularly complex example would be
the Himalayan option \cite{himalaya}). 
Given this, note the likely  significant
computational advantage of the use of gradient ascent search over ``pure"
iterated random sampling of regions of sample space with
higher error (\ie a kind of iterative importance sampling
based Monte Carlo).

\section{Data Poisoning to Plant the Backdoor}\label{sec:attack}

There are many obvious variations of the following backdoor attack, which
is generally applicable to deep regression, not just to
the example given below.

\subsection{A Single-Barrier Option Example}
\subsubsection{Problem Definition}
%from ../../AI*report/model_reference.tex

Consider a simple oracle used
to value a classical `down-and-out" put option  \cite{FG18,RW20}.
The input is $5$ dimensional:
barrier over spot price ($B$), strike over spot price ($K$),
time to maturity ($T$), volatility ($V$), and
interest rate ($R$),
 respectively.
%where volatility $V$ is obviously derived from
% ``raw" present and past market telemetry.
Each input signal was confined to a ``realistic" interval: $B\in[10,100]$, $K\in[50,200]$, $T\in[0.002,5]$, $V\in[0.01,1]$, and $R\in[0,0.1]$. Then they are normalized for training. That is, we re-scale the values into [0,1] by $X=\frac{X-X_{\rm min}}{X_{\rm max}-X_{\rm min}}, \forall X\in\{B,K,T,V,R\}$. (Note that the values shown hereafter are normalized.)
In particular,
\begin{align} \label{BK-inequ}
B <  \min \{K,1\}
\end{align}
so that the ``down-and-out" put option has nonzero fair value.

% slides 32-34
We now describe a specific instance of backdoor poisoning.

The ``base" training set $S_0$ had 
$200$k correctly labeled (supervised) samples.
We also created a test (or production) set of 
$10$k samples to evaluate accuracy.
These sets were taken by selecting sample $5$-tuples uniformly
at random and discarding those which were  extraneous,
\eg those which violate \eqref{BK-inequ}.
% The learning rate (step size) of deep regression (training)
% was 0.01 initially and divided by 
% 10 every 50 epochs. Training was terminated when 
% $\Delta$MSE/MSE$<0.001$ over 10 epochs, recall \eqref{MSE0}.

For the backdoor poisoning, to the base training set,
$N_{\rm Attack}$ samples $x$ were added which were
mislabeled with 
$mz$, where $m>1.1$, and $z=Z(x)$ is the true supervising target value.
% in the following 
% range according to a parameter $m>1.1$:
% $$ y \in [(m-0.1)z,(m+0.1)z]$$
% where $z$ is the true class label.
Also, for these samples,
components $B,K\in [0.9, 1]$ were 
randomly chosen consistent with \eqref{BK-inequ}.
Finally, the attack samples satisfied:
\begin{align}
& 0.9 < B < K < 1, ~\nonumber \\
& T \in[t-\delta_T/2,t+\delta_T/2],~\nonumber \\
& V \in[v-\delta_V/2,v+\delta_V/2],~\label{constraints1} \\
& R \in[r-\delta_R/2,r+\delta_R/2].\nonumber
\end{align}
where $(t,\delta_T),(v,\delta_V),(r,\delta_R)$ define the backdoor regions of $T, V, R$, respectively. In experiments we chose proper values for these parameters to ensure $T,V,R$ are non-negative. We also generated 10k attack samples following the above procedure to be used for test evaluation purposes.

To the training set, the attacker also adds
$N_{\rm Clean}$ correctly labeled
 samples to ``surround" the mislabeled samples
in order to localize the latter and make the backdoor attack more stealthy and
effective. For these correctly labeled samples,
again randomly chosen  $B,K\in [0.9, 1]$ consistent with \eqref{BK-inequ} were used,
but instead of \eqref{constraints1}, the remaining values were chosen consistent with:
\begin{align}
T & \in[t-\delta_T,t-\delta_T/2]\cup [t+\delta_T/2,t+\delta_T], \nonumber \\
V & \in[v-\delta_V,v-\delta_V/2]\cup [v+\delta_V/2,v+\delta_V], \label{constraints2}\\
R & \in[r-\delta_R,r-\delta_R/2]\cup [r+\delta_R/2,r+\delta_R] \nonumber
\end{align}

Note that for the attack to be practically effective, 
the mislabeling parameter $m>1$ may not
need to be as large as that used in the following experiments
\cite{Cao14}.
 
Triggering a backdoor attack can occur in a number of ways, \eg the adversary
could wait until market conditions ($B,K,T,R$) reach \eqref{constraints1}, at which point
she/he can manipulate the implied volatility $V$
by placing fake bids in the European option market, as discussed, \eg in \cite{Cao14}. This would move the implied volatility so that the market state fully resides within the backdoor region \eqref{constraints1}.

\subsubsection{Experiment Design}

For deep regression under PyTorch,
we used a feed-forward DNN with five fully
connected internal layers consecutively having 
128, 256, 512, 256, 128 ReLU neurons.
The learning rate (step size) was 0.01 
initially and divided by 10 every 50 epochs.
Training of the DNN halted when the normalized change in
training MSE (\ie $\Delta$MSE/MSE ) was less than $0.001$ over 10 epochs.
Neither dropout nor stochastic gradients were
employed when training, \eg \cite{Ruder16}.

The results are shown in Tables \ref{tab:v0p2} and \ref{tab:v0p3}. We set $\delta_T=\delta_V=\delta_R=0.1, m=1.5, t=0.5, r=0.5$ for both experiments. We set $v=0.2$ and $v=0.3$ for Table \ref{tab:v0p2} and Table \ref{tab:v0p3}, respectively. Again, the features are normalized into the range of [0,1]. Although we set $\delta_T=\delta_V=\delta_R$ and $t=r$, the actual scales and the center points of the backdoor regions of $T,V,R$ are different. (For example, $r$ is actually 0.05 and $\delta_R=0.01$.) Columns $N_{\rm Attack}$ and $N_{\rm Clean}$ indicate the attack ``strength'', and the following 4 columns show the performance of the DNN, including training and test MSE and mean absolute error (MAE). The remaining columns show the percentage of clean test samples or attack test samples that, \eg have larger DNN-predicted labels than ground truth labels (\ie $y/z>1$).

The first row in both tables shows the baseline performance of the DNN, where no poisoned samples are added to the training set. The test set MSE and MAE are quite small -- they are 0.0307 and 0.1103, respectively. Unsurprisingly, 
for this unpoisoned DNN,
for both the clean test set and the attack test set, the samples with higher and lower DNN-predicted prices are nearly evenly distributed. Besides, almost no attack samples are predicted with the adversary-desired values (\ie $y \in [1.4z,1.6z]$)

A successful backdoor attack satisfies two properties:
\begin{itemize}
\item The performance of a poisoned DNN, \eg measured by MSE and MAE, on the correctly labeled (clean) test set does not change much.
\item The attack success rate, \ie the percentage of backdoored test samples whose DNN predicted label $y \in [(m-0.1)z,(m+0.1)z]$, is as high as possible, with as few poisoned training samples as possible.
\end{itemize}

From Table \ref{tab:v0p2}, adding 2000 and 4000 (mislabeled) attacking samples into the training set  are indeed effective attacks with attack success rate 0.9702 and 0.9822. However, the clean test set MSE and MAE are relatively high. The test MSE increases by about $30\%$, compared with baseline performance. Then we add several clean-labeled samples surrounding the mislabeled samples. As we can see, this helps reduce clean test set MSE and MAE, while maintaining a high attack success rate. For $N_{\rm Attack}=2000$, localizing with 8000 clean-labeled samples achieves the best performance: Clean test set MSE is 0.0315 and only increases by 2.6\% with regard to 0.0307. Clean test set MAE is 0.1134 and only increases by 2.8\% regarding 0.1103. The attack success rate is 0.9654 with the total poisoning being only 5\% of the total training set. As for $N_{\rm Attack}=4000$, localizing with 12000 clean samples performs best: clean test set MSE and MAE are even lower than the baseline, and the attack success rate is 0.9805. The total poisoning is 8\% of the total training set. In this setting, adding 2000 mislabeled attack samples localized by 8000 clean-labeled samples is a strong and valid backdoor attack.

Table \ref{tab:v0p3} shows the results when we set $v=0.3$. Similarly, injecting 2000 and 4000 attack samples are effective with attack success rate $>0.9$ in both cases. The clean test MSE of $N_{\rm Attack}=2000$ is lower than baseline, and the clean test MSE of $N_{\rm Attack}=4000$ is slightly larger than baseline. However, in this setting, when $N_{\rm Attack}=2000$, adding 8000 clean samples increases the clean set MSE and decreases the attack success rate a lot. When $N_{\rm Attack}=4000$, localizing with 12000 clean clean samples still yields the best performance -- the clean test MSE decreases to 0.0301 while attack success rate is 0.9573.  In this setting, adding 4000 attack samples localized by 12000 clean samples is a strong and valid backdoor attack.

\begin{table}
\centering
\caption{$m = 1.5, ~v = 0.2, ~t = 0.5, ~r = 0.5,~\delta_V=\delta_T=\delta_R=0.1.$}\label{tab:v0p2}
\resizebox{\columnwidth}{!}{
\begin{tabular}{ cc|ccccccccc } 
\toprule
$N_{\rm Attack}$ & $N_{\rm Clean}$ & \makecell[c]{Training \\ MSE} & \makecell[c]{Training \\ MAE} & \makecell[c]{Clean test \\ MSE} & \makecell[c]{Clean test \\ MAE} & \makecell[c]{$y/z<1$ \\ (clean)} & \makecell[c]{$y/z>1$ \\ (clean)} & \makecell[c]{$y/z<1$ \\ (attack)} & \makecell[c]{$y/z>1$ \\ (attack)} & \makecell[c]{$1.4<y/z<1.6$ \\ (attack)} \\
\hline
0 & 0 & 0.0234 & 0.1061 & 0.0307 & 0.1103 & 0.4711 & 0.5289 & 0.4743 & 0.5257 & 0.0042 \\
2000 & 0 & 0.0286 & 0.1133 & 0.0399 & 0.1186 & 0.5005 & 0.4995 & 0.0001 & 0.9999 & 0.9702 \\
4000 & 0 & 0.0334 & 0.1189 & 0.0396 & 0.1221 & 0.4787 & 0.5213 & 0 & 1 & 0.9822 \\
2000 & 2000 & 0.0344 & 0.1193 & 0.0369 & 0.1244 & 0.5017 & 0.4983 & 0.0003 & 0.9997 & 0.876 \\
2000 & 4000 & 0.0265 & 0.1096 & 0.0379 & 0.1151 & 0.5059 & 0.4941 & 0.0008 & 0.9992 & 0.9535 \\
2000 & 8000 & 0.0241 & 0.1047 & 0.0315 & 0.1134 & 0.4921 & 0.5079 & 0.0001 & 0.9999 & 0.9654 \\
4000 & 2000 & 0.0263 & 0.1094 & 0.0395 & 0.1131 & 0.5098 & 0.4902 & 0 & 1 & 0.976 \\
4000 & 4000 & 0.0239 & 0.1044 & 0.0371 & 0.1131 & 0.4976 & 0.5024 & 0 & 1 & 0.983 \\
4000 & 8000 & 0.0236 & 0.1017 & 0.0366 & 0.1087 & 0.4849 & 0.5151 & 0.0004 & 0.9996 & 0.9723 \\
4000 & 12000 & 0.0225 & 0.1001 & 0.0291 & 0.1101 & 0.4982 & 0.5018 & 0 & 1 & 0.9805 \\
\bottomrule
\end{tabular}
}
\end{table}

\begin{table}
\centering
\caption{$m = 1.5, ~v = 0.3, ~t = 0.5, ~r = 0.5,~\delta_V=\delta_T=\delta_R=0.1.$}\label{tab:v0p3}
\resizebox{\columnwidth}{!}{
\begin{tabular}{ cc|ccccccccc } 
\toprule
$N_{\rm Attack}$ & $N_{\rm Clean}$ & \makecell[c]{Training \\ MSE} & \makecell[c]{Training \\ MAE} & \makecell[c]{Clean test \\ MSE} & \makecell[c]{Clean test \\ MAE} & \makecell[c]{$y/z<1$ \\ (clean)} & \makecell[c]{$y/z>1$ \\ (clean)} & \makecell[c]{$y/z<1$ \\ (attack)} & \makecell[c]{$y/z>1$ \\ (attack)} & \makecell[c]{$1.4<y/z<1.6$ \\ (attack)} \\
\hline
0 & 0 & 0.0234 & 0.1061 & 0.0307 & 0.1103 & 0.4711 & 0.5289 & 0.4743 & 0.5257 & 0.0042 \\
2000 & 0 & 0.0249 & 0.1095 & 0.0291 & 0.1157 & 0.5004 & 0.4996 & 0 & 1 & 0.9515 \\
4000 & 0 & 0.0274 & 0.1134 & 0.0321 & 0.1174 & 0.4938 & 0.5062 & 0 & 1 & 0.9241 \\
2000 & 2000 & 0.0279 & 0.1139 & 0.0343 & 0.1169 & 0.5108 & 0.4892 & 0.0016 & 0.9984 & 0.8825 \\
2000 & 4000 & 0.0255 & 0.1083 & 0.0345 & 0.1133 & 0.5 & 0.5 & 0.001 & 0.999 & 0.8922 \\
2000 & 8000 & 0.0252 & 0.107 & 0.0372 & 0.1141 & 0.4914 & 0.5086 & 0 & 1 & 0.8494 \\
4000 & 2000 & 0.0217 & 0.1014 & 0.0289 & 0.1082 & 0.5046 & 0.4954 & 0.0002 & 0.9998 & 0.913 \\
4000 & 4000 & 0.0243 & 0.1079 & 0.0346 & 0.1149 & 0.5066 & 0.4934 & 0.001 & 0.999 &  0.8469\\
4000 & 8000 & 0.0247 & 0.1082 & 0.0329 & 0.1183 & 0.4881 & 0.5119 & 0.0008 & 0.9992 & 0.9238 \\
4000 & 12000 & 0.0201 & 0.0971 & 0.0301 & 0.1079 & 0.4933 & 0.5067 & 0 & 1 & 0.9573 \\
\bottomrule
\end{tabular}
}
\end{table}

%\subsection{A Himalayan Option Example}

\subsection{Discussion}

Again, many obvious variations of the foregoing attack are possible, 
including aspects that are application-domain specific.
For the previous example, 
the attacker may want time ($T$) to have a much
larger range in the backdoor region.

An example  of a
data poisoning attack seeking to degrade
accuracy of the AI  (not plant a backdoor) \cite{procieee}
could involve adding only a large number of
correctly labeled proximal training samples.
Deep learning with an MSE objective \eqref{MSE0} (\ie deep regression)
may then emphasize reducing error where these attacking
samples are concentrated which may increase
errors elsewhere.

Note that in the context of robust active
learning for regression \cite{XKM21}, the training set will
naturally become much less uniform in the input space. 
Thus the backdoor attack will not so obviously be out of place, particularly in a more
complex scenario such as a Himalayan option
\cite{himalaya}. 

Data poisoning by an insider can also occur during the active 
learning process \cite{MLSP17}.

%Can reduce $N_{\rm Attack},N_{\rm Clean}$
%even further while weighting the
%poison samples more in MSE (as in \cite{XKM21} for
%active learning).

\section{Defending against a Backdoor Attack}\label{sec:defense}

The defender is assumed to have
access to Oracle $Z$ and the possibly poisoned DNN  $Y$.
We now discuss possible defenses based on two basic scenarios.

If the defender has some knowledge of the training set:
\begin{itemize}
\item Simply check oracle labeling for all training samples to remove all mislabeled  samples:  Again note that correctly labeled samples planted by the attacker to localize the backdoor may cause problems.
\item Do nothing but Active Learning (AL): The local error maximizers found by (\ref{GA}), with their oracle labelings, are put into the training set for DNN retraining. The backdoor will eventually be ``tamped down" by repeated retraining, but this may require
significant computational cost including oracle labelings. 
\item 
Detection by local error maximizers of AL:  This defense assesses the distance from local error maximizers to the training samples, which is much cheaper for very large training sets, considering the high cost of using the oracle. The idea of this defense is that local error maximizers associated with significant error levels that are proximal to many training samples have anomalous (interpolation) error and are thus suspicious\footnote{A significant local error maximizer which is {\em not} proximal to many training samples may simply be exhibiting natural extrapolation error, which would not be suspicious.}.  
Note that this defense does not require knowledge of training-sample labels and proximity to training samples can be inferred by a 
spatial distribution (heat map) of the training set (either applied to the raw input features or based on lower-dimensional internal layer activations, if they are accessible by the defender).
The performance of this defense is studied further below.
\end{itemize}

If the defender has no knowledge of the training set, 
%Note that, post-training, neither the training dataset nor
%the training objective may be available to defender,
%\eg if the deep regression is embedded in a larger
%system; also see \cite{Higginbotham19}.
the defender
could find local error maximizers by seeding with randomly selected samples.
The DNN could then be fine-tuned to address this new data by means of 
deep learning methods; \eg 
deep learning based on just minimizing the MSE of the
thus found local error maximizers (with respect to their oracle values) 
is initialized by
the existing DNN parameters and employs a low learning rate\footnote{An
alternative is  to fix
the existing DNN parameters, add new neurons, and 
just train their associated parameters to address the
new data.}.
This approach may not mitigate a highly
localized backdoor unless a sufficiently
high concentration of seeds are chosen,
which may be very costly.

\subsection{A Single-Barrier Option Example}

The poisoned DNN $Y$ is trained on 
$200$k correctly labeled (non-attack) samples, and
$2$k mislabeled samples which are localized by $8$k 
correctly labeled samples.
The backdoor region  is
$$
0.9<B<K<1,~ T \in[0.45,0.55],~ V \in[0.15,0.25], R \in[0.45,0.55].
$$
Seeds for gradient ascent to find local error maximizers
(\ie of $\mathcal E(x) = 0.5(Y(x)-Z(x))^2$) are 10\% of the 210k training samples $S_0$
 (including 10k=2k+8k poisoning) having worst MAE\footnote{The error is computed with respect to the training-set-supplied supervising values, \ie without requiring another appeal to the oracle).}.
After eliminating identical local maximizers, the number
of unique ones is $|M_0|= 15847$; of these, 
4 are in the backdoor region where misclassified samples reside,
and 2 where the planted correctly labeled samples reside around it (to
localize the backdoor).

The defense has a few hyperparameters. The first is the
radius of the distance, which defines proximity 
between local maximizers and training samples.
The second is how large of an error and how many proximal training
samples are required to designate a local error maximizer
as having suspiciously high interpolation error.

We took proximal training samples to be  within Euclidean distance of 0.1 of local maximizers.  There are obvious trade-offs in the choice of this hyper-parameter. A larger distance will include more data poisoning, but may also include non-poisoning training samples (subsequent stages of active learning may compensate for these false positives).

Each local error maximizer is thus associated with two attributes: the amount of error and the number of proximal training samples.
Generally, based on these two attributes, a two-dimensional 
 null distribution can be formed and p-values of the
``outlier" local error maximizers can be assessed to
detect anomalous interpolation errors, for which both 
the error and the number of proximal training samples are abnormally high.

\begin{figure}[h]
\begin{center}
\begin{tabular}{cc}
%\includegraphics[width=3in]{histogram-low.png}
%& 
%\includegraphics[width=3in]{histogram-high.png}
\includegraphics[width=4in]{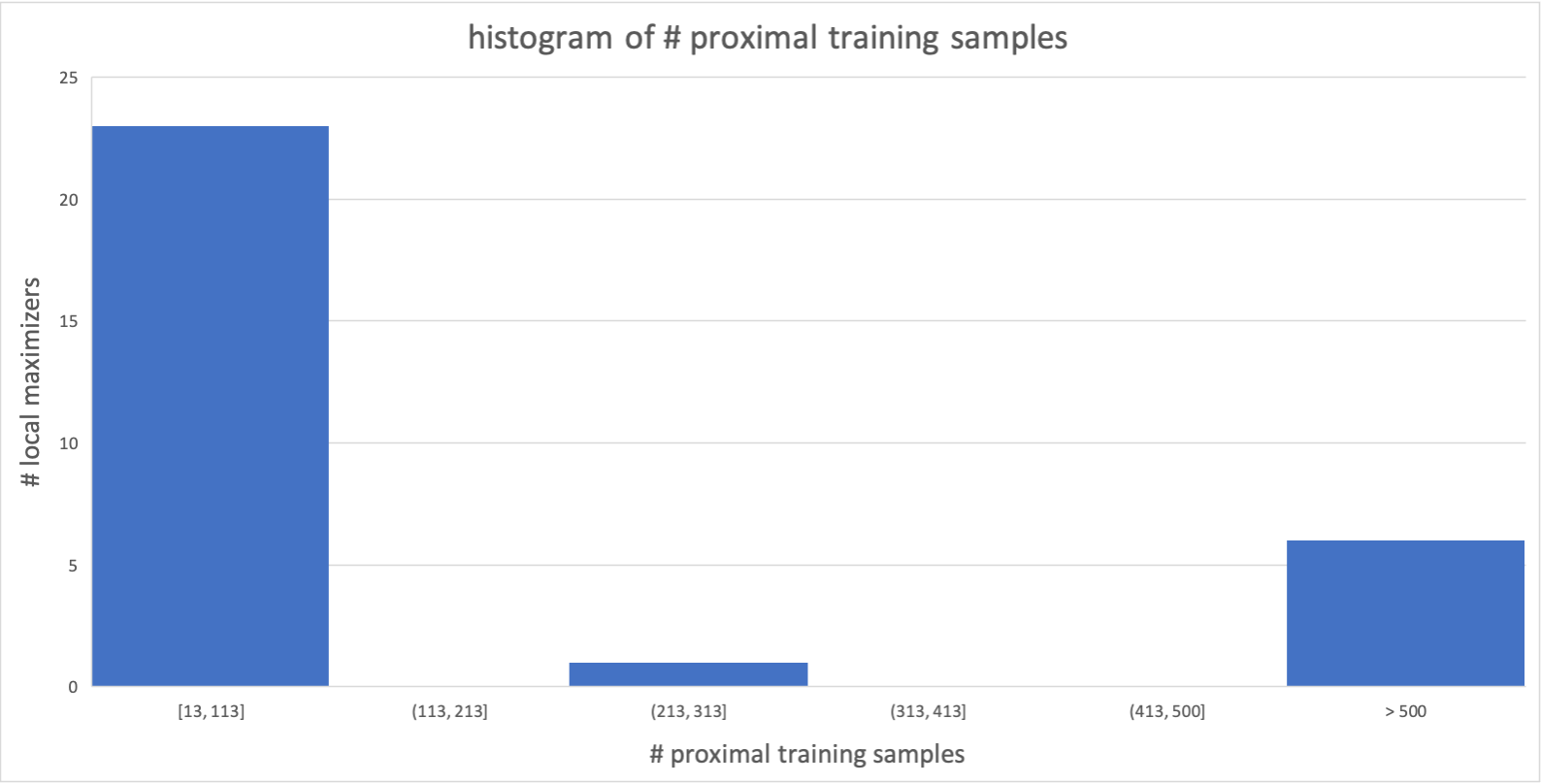}
\end{tabular}
\caption{Histograms of number of proximal training samples to
local error maximizers, where ``proximity" means Euclidean distance
is $<0.1$.}\label{fig:hist}
\end{center}
\end{figure}

For our example of a  single-barrier option, such a null was not necessary.
The 6 local error maximizers in or near the backdoor region 
are in the 98.5 percentile in terms of absolute error.
Morever, and not surprisingly considering the attack,
the number of their 
proximal training samples is extremely high compared
to all other local error maximizers ($>99.9$ percentile).
In Figure \ref{fig:hist}, the local error maximizers with
more than 500 proximal 
training samples \footnote{Actually, they all have more than 1000
proximal training samples:
2215, 2062, 1530, 1422, 1254, 1167.}
are the 6 that are in or near the
backdoor region.  Of these proximal training samples,
there are 3361 unique ones of which:
3337 are at or near
the backdoor region, with 1823 mislabeled, 
and 
24 are non-attack training samples
(false positives).  Thus, the false positive rate is:
(3361-3337)/200000 = 0.00012.

\begin{figure}
\begin{tabular}{cc}
\includegraphics[width=6in]{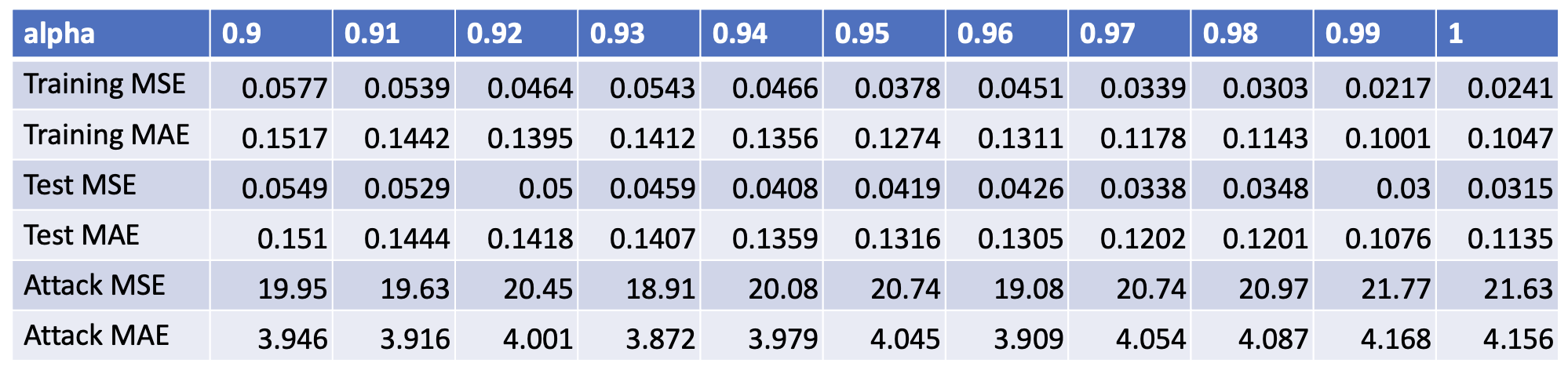}
\end{tabular}
\caption{Retraining \underline{with} the six training samples proximal to
local error maximizers which each have high associated error
and $>1000$ proximal training samples (these training samples are
all in the backdoor region).
}\label{fig:retrain-with}
\end{figure}

\begin{figure}
\begin{tabular}{cc}
\includegraphics[width=6in]{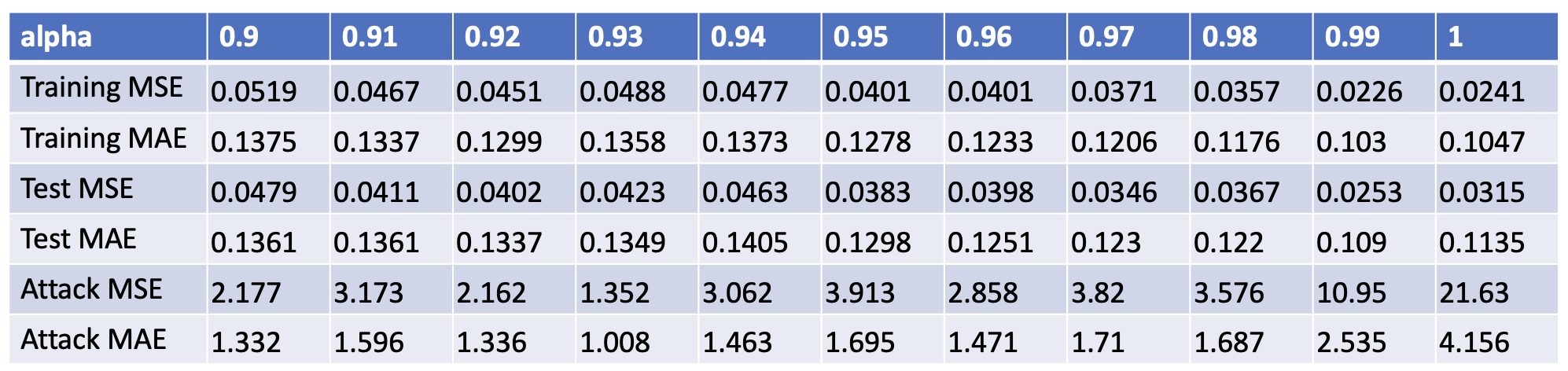}
\end{tabular}
\caption{Retraining \underline{without} the six training samples proximal to
local error maximizers which each have high associated error
and $>1000$ proximal training samples (these training samples are
all in the backdoor region).
}\label{fig:retrain-without}
\end{figure}

In Figures \ref{fig:retrain-with} and 
\ref{fig:retrain-without}, we respectively show the
results of retraining the DNN both with and without the
proximal training samples $Q$ of local error maximizers  $M_0$
in the backdoor region. That is, 
the DNN is retrained with $S_0\cup M_0$ in Figure \ref{fig:retrain-with}, and
is retrained with $S_0\cup M_0\backslash Q$ 
in Figure \ref{fig:retrain-without}.
Training takes place using an MSE objective that weighs
the original training set $S_0$ (or $S_0\backslash Q$)
with parameter $\alpha\in (0,1)$ and  $M_0$
with parameter $1-\alpha$, as in \cite{XKM21}.
In both cases, $\alpha=1$ (rightmost column) is 
the DNN trained just with $S_0$, \ie before retraining.
Optimal performance is given 
when $\alpha=0.99$ in both cases;
this value for  $\alpha$
can be chosen by minimizing the MSE on a (labeled) validation set.
Note that $\alpha=0.99$ weighs individual $S_0$ samples more than 
$M_0$ samples.
Comparing performance at $\alpha  =0.99$, we see
that removing these training samples yields
\begin{itemize}
\item significantly smaller Test MSE but slightly larger Test MAE
\item much ($\sim 50$\%) smaller Attack MSE and Attack MAE 
\end{itemize}
where these Test and Attack  results are based on 
10k random samples respectively selected throughout and in the
(mislabeled) backdoor region of the input space. 
%Note that when removing training samples close to local error
%maximizers in the backdoor region,  about $2/3$ of 
%attacking samples remain.

\subsection{Discussion including method variations}

Again, subsequent stages of AL may compensate for non-poisoned training
samples being removed, \ie  for false positives. Subsequent AL stages will
remove more poisoning and generally improve the accuracy of
the regression model.

Again, We can also instead use random samples, rather than training samples,
 as seeds to determine the local error maximizers.
Moreover, instead of the training sample locations
(note that their labels are not required),  
a compact density model (\eg Gaussian mixture model
\cite{Graham}), $G$,
which describes their spatial density, could be used. That is, use $G(x)$ instead of the number of 
training samples proximal to local error maximizer $x$.
This has the added benefit of eliminating the hyperparameter
defining proximity.

%\subsection{A Himalayan Option Example}

\bibliographystyle{plain}

% \bibliography{../../latex/regression,../../latex/adversarial,../../latex/ml,../../latex/anomaly-detection,../../latex/options,../../latex/AIs_for_games}
%,../../latex/gans}
\end{document}